\documentclass[10pt,twocolumn,letterpaper]{article}

\usepackage[pagenumbers]{cvpr} % To force page numbers, e.g. for an arXiv version

\usepackage[dvipsnames]{xcolor}

\usepackage{enumitem} % reduce space in enum
\usepackage{booktabs}
\usepackage{multirow}
\usepackage{rotating}
\usepackage{adjustbox}
\def\eg{\emph{e.g}\onedot}

\def\ie{\emph{i.e}\onedot}

\usepackage{mathtools}
\usepackage{textcomp}
\usepackage{gensymb}
\usepackage{amsmath}
\usepackage{soul}
\usepackage{bbm}
\usepackage{graphicx}
\usepackage{tikz}
\usetikzlibrary{calc}
\usepackage{comment}
\usepackage[dvipsnames]{xcolor}
\usepackage[accsupp]{axessibility}

\newcommand{\myparagraph}[1]{\vspace{4pt}\noindent\textbf{#1.}}

\definecolor{cvprblue}{rgb}{0.21,0.49,0.74}
\usepackage[pagebackref,breaklinks,colorlinks,citecolor=cvprblue]{hyperref}

\def\paperID{8} % *** Enter the Paper ID here
\def\confName{CVPR}
\def\confYear{2024}

\title{EarthMatch: Iterative Coregistration for Fine-grained Localization \\ of Astronaut Photography}

\author{Gabriele Berton$^{1}$
\quad
Gabriele Goletto$^{1}$
\quad
Gabriele Trivigno$^{1}$
\\
\quad
Alex Stoken$^{2}$
\quad
Barbara Caputo$^{1}$
\quad
Carlo Masone$^{1}$\\
$^{1}$Politecnico di Torino 
$^{2}$Jacobs Technology, NASA Johnson Space Center\\
}

\begin{document}
\maketitle

\begin{abstract}
Precise, pixel-wise geolocalization of astronaut photography is critical to unlocking the potential of this unique type of remotely sensed Earth data, particularly for its use in disaster management and climate change research. Recent works have established the Astronaut Photography Localization task, but have either proved too costly for mass deployment or generated too coarse a localization. Thus, we present EarthMatch, an iterative homography estimation method that produces fine-grained localization of astronaut photographs while maintaining an emphasis on speed. We refocus the astronaut photography benchmark, AIMS, on the geolocalization task itself, and prove our method's efficacy on this dataset. In addition, we offer a new, fair method for image matcher comparison, and an extensive evaluation of different matching models within our localization pipeline. Our method will enable fast and accurate localization of the 4.5 million and growing collection of astronaut photography of Earth.
Webpage with code and data at 
{\small{\url{https://earthloc-and-earthmatch.github.io}}}

\end{abstract}

\section{Introduction}
\label{sec:introduction}
Computer vision plays a pivotal role in analyzing remotely sensed Earth observations, demonstrating efficacy across diverse applications including Earth monitoring, atmospheric and climate research, and emergency response strategies \cite{ghaffarian2021effect, fisher_issrd, dandois2010remote, camps2021deep}.
Much of this remotely sensed imagery comes from satellites devoted to Earth observations, but a peculiar, yet valuable complementary source of data is astronaut photography. 
More than 4.5 million photos of Earth\footnote{\url{https://eol.jsc.nasa.gov/}} have been taken by astronauts, primarily from the International Space Station (ISS) during its over 20 years of continuous operation in low Earth orbit (roughly 400 km), a vantage point which offers a privileged perspective with a field of view that can span up to thousands of kilometers.

\begin{figure}[t]
    \begin{center}
    \includegraphics[width=0.99\linewidth]{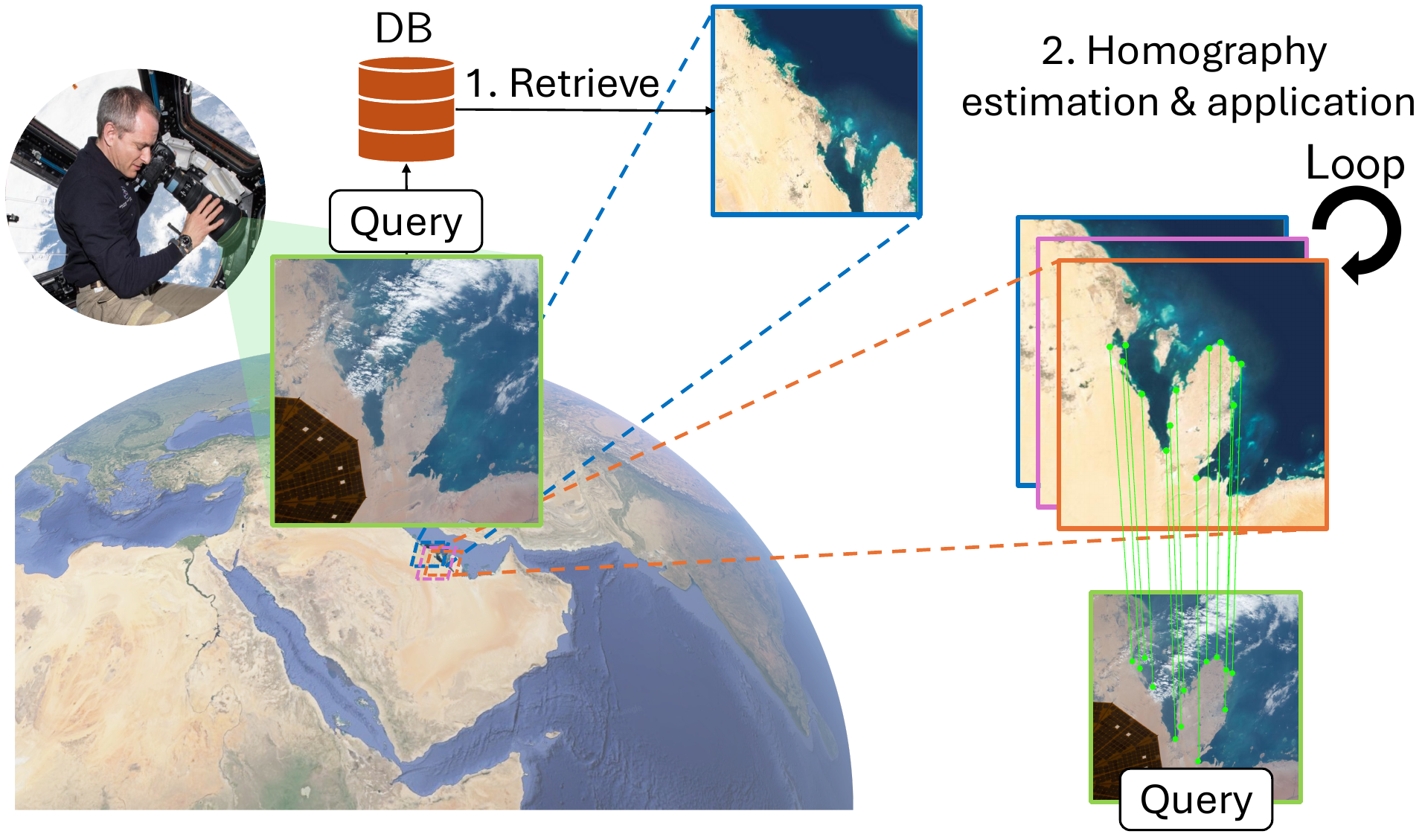}
    \end{center}
    \vspace{-3mm}
    \caption{\textbf{EarthMatch}: 
    To produce a pixel-wise geolocalization of an astronaut photograph (the \emph{query}), we first retrieve a possible candidate from a worldwide database of satellite images.
    We then compute image correspondences and coregister the two images iteratively, yielding a precise query localization and confidence value.
    }
    \vspace{-3mm}
    \label{fig:teaser}
\end{figure}

Astronaut photography is particularly valuable due to its unique characteristics. For instance, the orbital speed and trajectory of the ISS enables quick response time in case of natural disasters and other emergencies, as astronauts can be promptly alerted and take timely photographs which, after manual localization, are provided to first responders on the ground \cite{IDC_Stefanov}. Protocols based on this concept have been successfully implemented during the 2013 Haiyan's cyclone, as well as to handle flooding events, wildfires, and several other crises throughout the planet~\cite{Stefanov_2015_disaster, elliott2020earth, miyazaki2015reviews}. Geolocated astronaut photographs have also been used for research purposes across a variety of Earth science topics \cite{DIEGO_thermal_mission, wilkinson_gunnell_2023, astro_photos_climate_patterns, TLEs}.
Furthermore, unlike unmanned satellites, behind each photo is a skilled human (as astronauts undergo photography and geography training) that interprets the observable landscape and is not constrained to a fixed perspective (see \cref{fig:teaser}). Adjusting parameters like orientation, focal length and viewing angle offers intriguing possibilities for data collection which are difficult or impossible to achieve with conventional satellites. 
Yet, the very same human-in-the-loop nature of astronaut photographs also makes their precise localization challenging, since only a coarse estimate can be inferred from the location of the ISS at the photo acquisition time, and manual verification has to be performed on each image for precise localization.
Despite the high manual localization cost, over 300k (\ie still less than 10\%) of images have been manually localized, demonstrating the importance and value of geotagged astronaut photographs.
In order to unlock the full potential utility of this vast corpus of images, an automated fine-grained geolocalization  solution is needed - a task named Astronaut Photo Localization (\textbf{APL}) by EarthLoc~\cite{Berton_2024_EarthLoc}. 

\begin{figure}
    \centering
    \begin{tikzpicture}[every node/.style={inner sep=0pt}]
        \node (img1) at (0,0) {\includegraphics[width=0.30\linewidth]{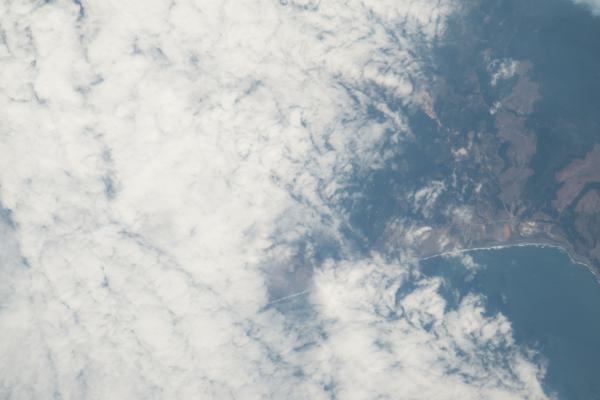}};
        \node (img2) at (0.33\linewidth,0) {\includegraphics[width=0.30\linewidth]{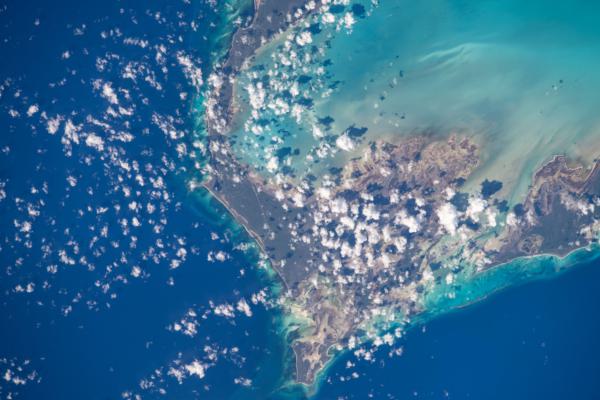}};
        \node (img3) at (0.66\linewidth,0) {\includegraphics[width=0.30\linewidth]{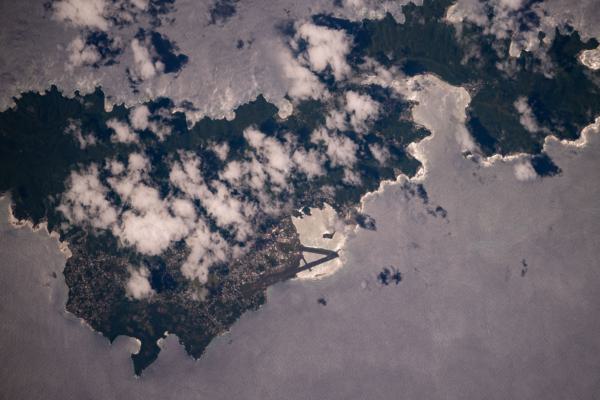}};
        \node (img4) at (0,-0.35\linewidth) {\includegraphics[width=0.30\linewidth]{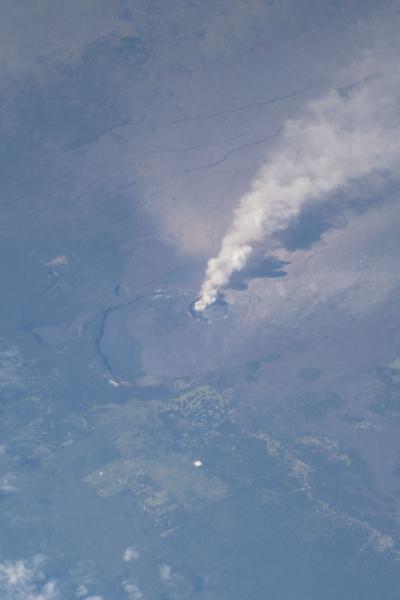}};
        \node (img5) at (0.33\linewidth,-0.35\linewidth) {\includegraphics[width=0.30\linewidth]{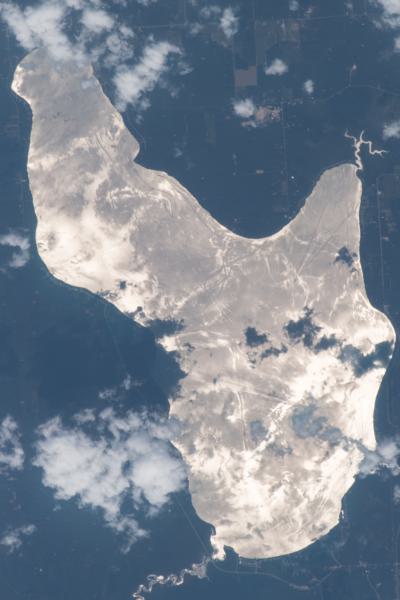}};
        \node (img6) at (0.66\linewidth,-0.35\linewidth) {\includegraphics[width=0.30\linewidth]{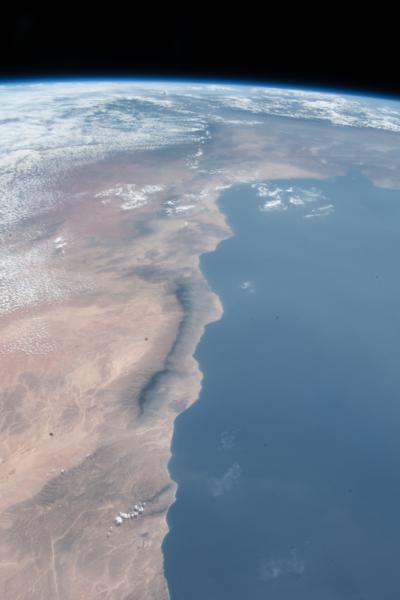}};
        
        \draw[red, line width=2pt] (img6.south west) ++(0.02,0.04) rectangle ($(img6.north east) + (-0.04,-0.04)$);
    \end{tikzpicture}
    \vspace{-3mm}
    \caption{\textbf{Examples of astronaut photo queries} from the AIMS dataset \cite{Stoken_2023_FMAP} which we use in our experiments.
    The rightmost bottom image is an example of a photo with high tilt/oblique, which we remove prior to our benchmark evaluation. These images are also the least useful/informative for Earth science researchers.}
    \vspace{-3mm}
    \label{fig:ex_queries}
\end{figure}

To this end, the work of Find My Astronaut Photo (FMAP)~\cite{Stoken_2023_FMAP} proposed the first localization solution for this problem and released a labelled dataset which is of paramount importance for future works aiming to assess the reliability of methods before they are applied to unlabelled images.
FMAP uses a brute-force approach based on pairwise matching against a reference obtained from satellite data. The high latency of this extensive matching makes it unsuitable for applications where speed is of the utmost importance, such as assisting in natural disaster response. 
Recently, EarthLoc~\cite{Berton_2024_EarthLoc} showed that retrieval methods can drastically reduce the uncertainty on the area of interest by identifying a handful of candidate regions which are most similar to the query image to be localized. Candidates are retrieved from a worldwide database of cloudless Sentinel-2 satellite images\footnote{\url{https://s2maps.eu/}},
which divide the Earth into regular, rectilinear tiled regions.  
Despite its robust performance, a pipeline based solely on image retrieval does not provide pixel-wise localization of the query, nor does it provide a confidence measure for its predictions.

In light of these considerations, we advocate for a hierarchical pipeline that combines an EarthLoc-like retrieval step with a coregistration method capable of providing an accurate location estimate as well as a reliable confidence score that allows for the rejection of false positives.
To this end, we present EarthMatch, a simple yet efficient fine-grained geolocalization algorithm based on iterative homography refinement, and demonstrate its effectiveness through a comprehensive benchmark.
In summary, our main contributions are:
\begin{itemize}
    \item an APL pipeline made of (i) a retrieval model to obtain possible candidate locations for a given query, and (ii) EarthMatch to confidently obtain a single final prediction through an iterative coregistration algorithm;
    \item a thorough benchmark in which we implement our algorithm with a large number of existing image matching methods, ranging from sparse detectors, learned matchers, handcrafted methods as well as dense warp estimators;
    \item an extension to the Astronaut Imagery Matching Subset (AIMS) dataset \cite{Stoken_2023_FMAP} with 268 astronaut photographs and their the top-10 predictions (satellite images) from the current SOTA retrieval APL model \cite{Berton_2024_EarthLoc}, and we release this extended dataset to foster future research.
\end{itemize}
The code to run the benchmark, as well as scripts to download the data, is available at 
{\small{\url{https://EarthLoc-and-EarthMatch.github.io/}}}.
All the images localized with EarthMatch (plus images localized with FMAP) can be explored in a convenient interactive format at 
{\small{\url{https://eol.jsc.nasa.gov/ExplorePhotos/}}}.

\section{Related work}
\label{sec:related_work}

\myparagraph{Feature detection and description}
Finding salient points in an image, and their associated descriptors, is a cornerstone of computer vision. Points and descriptions are employed for a variety of different tasks, such as Visual Localization \cite{Sattler_2018_aachen_daynight, sarlin_2019_coarse}, Structure from Motion (SfM) \cite{schoenberger_2016_sfm, schoenberger_2016_mvs} and Simultaneous Localization and Mapping (SLAM) pipelines \cite{durrant_2006_simultaneous, bailey_2006_simultaneous2, artal_2015_orb}.
Classical methods rely on a detect-then-describe approach, typically based on handcrafted descriptors computed from local derivatives of the image~\cite{Lowe_2004_sift}. The most popular among these methods is SIFT~\cite{Lowe_2004_sift} which provides a framework for feature detection and description that is robust to scale and distortions. Follow-up works, such as SURF~\cite{Bay_2008_surf}, focused on improving efficiency, while others like ORB \cite{artal_2015_orb} combine fast keypoint detection with a robust binary descriptor.

As deep learning gained prominence, learnable approaches for both detection and description were introduced. In particular, \cite{serra_2015_desc, tian_2017_l2net} are the first to popularize sparse keypoint detection and description with deep neural networks, relying on contrastive strategies to learn patch-level descriptors with a CNN.
Later, SuperPoint~\cite{DeTone_2018_superpoint} introduced a self-supervised framework for training on synthetic shapes with arbitrarily defined keypoints.
Conversely, D2-Net~\cite{Dusmanu_2019_D2Net} and R2D2 \cite{Revaud_2019_r2d2} propose a framework to jointly detect and describe features, modeling keypoints as local maxima of the feature maps.
Other subsequent works define salient points as those that maximize the matching probability, again trained through self-supervision \cite{Gleize_2023_ICCV,Tyszkiewicz_2020_disk}.

Among the most recent solutions, ALIKE \cite{Zhao_2023_alike} uses a novel detection module based on a patch-wise softmax relaxation and has been further extended in ALIKED \cite{Zhao_2023_aliked} by exploiting deformable convolutions that adapt to the keypoints' support. 
DeDoDe~\cite{Edstedt_2023_dedode} decouples the detection and description steps into two independent models; Steerers~\cite{Bokman_2023_steerers} further improves upon this approach with a formulation designed to be rotation invariant.

\myparagraph{Image matching}
Our task of interest, localizing astronaut photography, has challenges akin to the problem of wide baseline image matching \cite{jin_2021_image}. This is mainly due to the substantial appearance shift with respect to our reference set and the minimal overlap that we typically face between a query and its retrieved candidate. Image matching tries to find correspondences (\emph{e.g.}, points or regions) among images by comparing features. 
Traditionally, matching was performed on handcrafted descriptions with a nearest neighbor search over descriptors followed by Lowe's ratio test~\cite{Lowe_2004_sift}, or via mutual nearest neighbor in the case of learnable, sparse methods~\cite{Revaud_2019_r2d2}.
The landscape evolved with SuperGlue~\cite{Sarlin_2020_superglue}, a graph neural network-based approach for matching features, which leverages attention mechanisms to deal with the challenges of significant viewpoint change or occlusion. The introduction of LoFTR~\cite{Sun_2021_loftr} represented a departure from discrete feature detection towards a detector-free, semi-dense matching paradigm, using transformers to coarsely match features. These methods stand out for their effectiveness even in texture-sparse scenarios. Following LoFTR, several other methods followed a similar approach~\cite{Wang_2022_matchformer, Tang_2022_quadtree, Huang_2023_Adaptive, Chen_2022_aspanformer, Bokman_2022_se2loftr,Zhou_2021_patch2pix}.
An alternative line of research focuses on dense warp estimation between image pairs, from which keypoints can be sparsely sampled when needed~\cite{Edstedt_2023_dkm}. Recently, RoMA~\cite{Edstedt_2023_roma} proposed to exploit a universal vision foundation model, DINOv2~\cite{Oquab_2023_dinov2}, for the task of dense feature matching; given the coarse nature of vision transformers, the authors combine it with a specialized CNN for match refinement, achieving state-of-the-art results on a variety of downstream benchmarks~\cite{jin_2021_image, Li_2018_megadepth, Balntas_2017_hpatches}.
In this work, we are interested in using pairwise matchers to asses the similarity of two images. Typically these methods are benchmarked on downstream applications like homography or pose estimation; however, in our case we match images which may exhibit only partial spatial overlap or remain entirely distinct. Thus, our goal is to understand whether they depict the same place or not.
Hence, similar to what is done in~\cite{Stoken_2023_FMAP}, we use the keypoints resulting from matching as a similarity measure. Additionally, we also use these salient points to estimate a projective transformation between the images.
Iterative homography estimation itself has been studied as a technique to refine the co-registration between two images. Traditionally, these methods were based on the Lucas-Kanade algorithm \cite{baker2004kanade,lucas81iterative}, which relies on photometric error. In \cite{Cao_2022_iterative_homography}, the authors propose an end-to-end trainable network for deep iterative homography estimation. Recently, \cite{Bellavia_2023_refinementIM, Bellavia_2024_bare_homography} proposed novel handcrafted techniques for match and homography refinement.

\myparagraph{Astronaut photography localization}
Recent literature has shown a growing interest in the problem of localizing photographs from astronauts. Pioneering works Find My Astronaut Photo (FMAP)~\cite{Stoken_2023_FMAP} and \cite{schwind_nighttime_georef} both rely on pairwise comparison between queries and reference data. While the former focuses on daytime pictures with satellite imagery as reference, the latter tries to localize night-time images by generating synthetic data to bridge the domain gap.
While showing convincing performance, these methods are hindered by the intrinsic complexity of matching images without prior knowledge of candidate regions. Recently, EarthLoc~\cite{Berton_2024_EarthLoc} introduced a retrieval method designed to handle the specific challenges of astronaut photography, namely large orientation, scale and appearance variations. Additionally, astronaut photos were used as a downstream benchmark to demonstrate a rotation equivariant detector proposed by Steerers~\cite{Bokman_2023_steerers}.  

In this work, we propose to exploit EarthLoc's candidate predictions to reduce the search area toward a few likely regions. Given EarthLoc top-k predictions, we apply an iterative matching algorithm that is able to correctly and precisely localize the query when a suitable candidate is present. Furthermore, our method provides a confidence metric, enabling the identification of instances where a candidate lacks overlap with the query, thereby establishing a method to reject false positive candidates from retrieval.

\begin{figure*}[ht!]
    \begin{center}
    \includegraphics[width=0.99\linewidth]{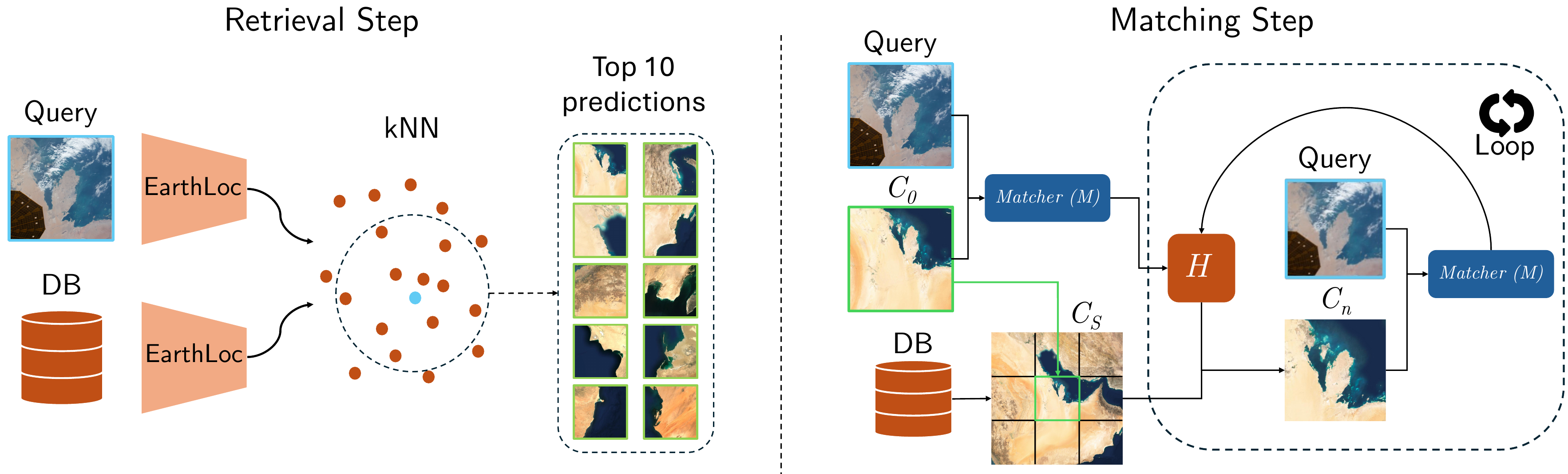}
    \end{center}
    \vspace{-4mm}
    \caption{\textbf{Left: Overview of retrieval step}, which, for a given query, retrieves candidates/predictions from a worldwide database of geo-tagged images.
    \textbf{Right: Overview of matching step.} The matching pipeline takes as input the query and a retrieved candidate. Surroundings of the candidate are obtained from the database, and then the iterative coregistration (in the form of matching and homographic transformation) is performed.}
    \vspace{-3mm}
    \label{fig:overview_method}
\end{figure*}

\section{EarthMatch}
\label{sec:pipeline}

\myparagraph{Overview}
We propose a hierarchical pipeline to solve the APL task and precisely localize astronaut photographs.
Our approach is inspired by the re-ranking approaches commonly adopted in Image Retrieval~\cite{Noh_2017_delf,Cao_2020_delg} and Visual Place Recognition~\cite{Zhu_2023_r2former, Hausler_2021_patch_netvlad, Barbarani_2023_local_features_benchmark, Lu_2024_SelaVPR}, where a fast retrieval method is used first to obtain a shortlist of candidates, followed by a more computationally demanding matching step to refine the estimate.

An overview of the pipeline is presented in \cref{fig:overview_method}: we rely on an APL image retrieval model (e.g. EarthLoc \cite{Berton_2024_EarthLoc}) which, for a given query, provides a shortlist of candidates.
Such candidates are retrieved from a worldwide database of geotagged satellite images (\ie images for which the pixel-wise location, in the form of geographic coordinates, is known).
Subsequently, to refine these candidates, we re-cast the matching step in the form of an iterative coregistration algorithm, called \textbf{EarthMatch}, that is tailored for the task of APL (see \cref{fig:overview_method}).
EarthMatch takes as input a query image $Q$ (\ie, the astronaut photograph) and a candidate image $C$ (\ie, a satellite image), where the candidate has been proposed by the retrieval method as potentially overlapping the query.
The algorithm's goal is twofold: (1) understand if the two images have any overlap and (2), if there is an overlap, precisely estimate the overlap, in the form of a homography matrix.
A precise estimate of the overlap allows extraction of the pixel-wise position of the query, given the known geographic boundaries of the candidate.
To achieve this goal, and to deal with the large baselines that retrieval candidates present, we iteratively refine the homography estimate.

\myparagraph{Problem setting}
The hierarchical pipeline takes as input an RGB query $Q$, and through EarthLoc~\cite{Berton_2024_EarthLoc} we retrieve a set of candidates. Candidates are extracted from a reference database that covers the landmass of the entire planet, as detailed in \cref{sec:dataset}.
We start from a RGB candidate $C_0$ with its corresponding \emph{footprint} $F_C$, defined as the latitude and longitude of its four corners, \ie, $F_C = \{ (lat_0, lon_0), (lat_1, lon_1), (lat_2, lon_2), (lat_3, lon_3) \}$ and apply our EarthMatch iterative coregistration algorithm.
The goal is to compute a homography $H_Q$ that maps the candidate onto the query such that they overlap, thus providing the footprint of the query, $F_Q$.

\subsection{Single-step pipeline}
The simplest way to obtain $F_Q$ is via a single homography estimated between the query $Q$ and the initial candidate, $C_0$. 
Such a pipeline requires a matcher $M$ which, given $Q$ and $C_0$, produces an \textit{overlap confidence}, a value that indicates the likelihood of the two images overlapping, and the homography $H_0$ to map the pixels of $C_0$ onto $Q$. Formally, $x_Q = H_0 x_{C_0}$, where $x_Q, x_{C_0}$ are expressed in homogeneous coordinates.
The homography $H_0$ is then simply used to estimate $F_Q$ given the footprint of the candidate $F_C$.

Although this single-step pipeline would be sufficient in the ideal case - with a perfect matcher - in practice matchers have a hard time producing enough well-distrubuted keypoint correspondences to estimate a good homography. This is especially true when the overlap is limited, or when the transformation is significant, for example those that include large degrees of rotation and re-scaling. Further, aligning Earth observations imagery can have added difficulty, as imagery often depict barren, featureless landscapes which are notoriously hard to match \cite{Ma_2020_survey}. In cases when there are many correspondences, but they are poorly distributed or, worse yet, highly localized to a single region of $Q$, $H_0$ is poorly constrained, resulting in a transform that does not well model the entire image area.

To overcome these issues, we implement a multi-step, iterative pipeline, which reduces the burden on the matcher by iteratively producing candidates that have increasingly more overlap with the query.

\subsection{Iterative pipeline}
The iterative pipeline picks up from the end of the single-step version, with a homography $H_0$ estimated from $Q$ and $C_0$.  From here, we apply $H_0$ to $C_0$ in order to obtain a new image $C_1$, which by construction will have a larger overlap with the query. Then, we use our new $C_1$ as a matching target for $Q$ and obtain a second, improved estimate of the homography $H_1$.

\begin{figure}
    \begin{center}
    \includegraphics[width=\linewidth]{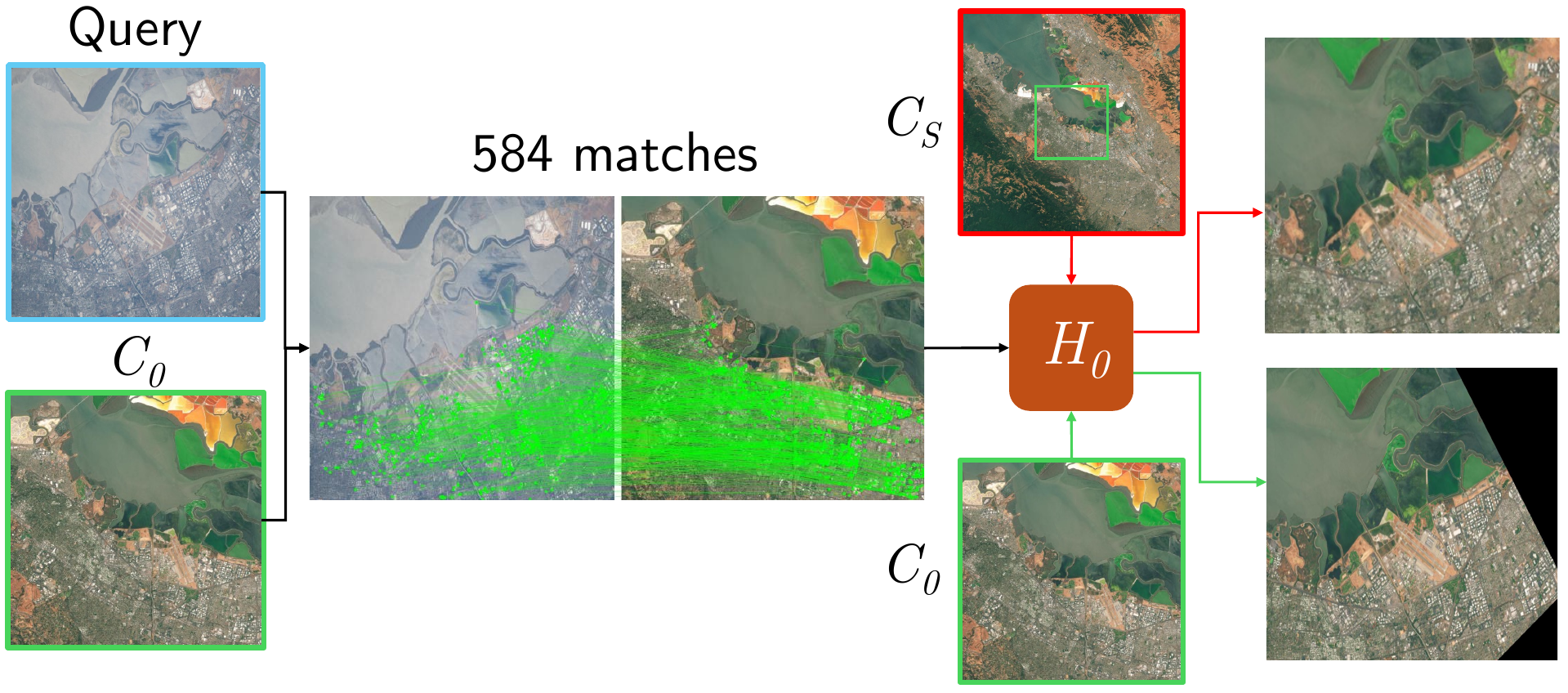}
    \end{center}
    \caption{\textbf{Coregistration example.} Directly applying the homography to the candidate image results in empty areas (see bottom-right image). Transforming the image along with its surroundings solves this issue (top-right image).
    }
    \label{fig:coregistration_example}
\end{figure}

Applying the homography directly to $C_0$ can lead $C_1$ to have some ``empty'' areas from locations depicted in $Q$ but not in $C_0$. These areas require filling, often handled with black padding in computer vision libraries.
To minimize this issue, instead of applying the homography directly to $C_0$, we apply it on a ``wider'' satellite image (an image of a larger area), comprising $C_0$ and its 8 adjacent tiles, which we call $C_S$, where the $S$ stands for \emph{surroundings}.
This reduces the chance that $C_1$ will have ``empty'' areas, unless a very strong homography is applied, which in practice we found to happen in a negligible number of cases.
Examples of this process, by applying the homography to $C_0$ versus applying it to $C_S$, is shown in \cref{fig:coregistration_example}.

\begin{figure*}
    \begin{center}
    \includegraphics[width=\linewidth]{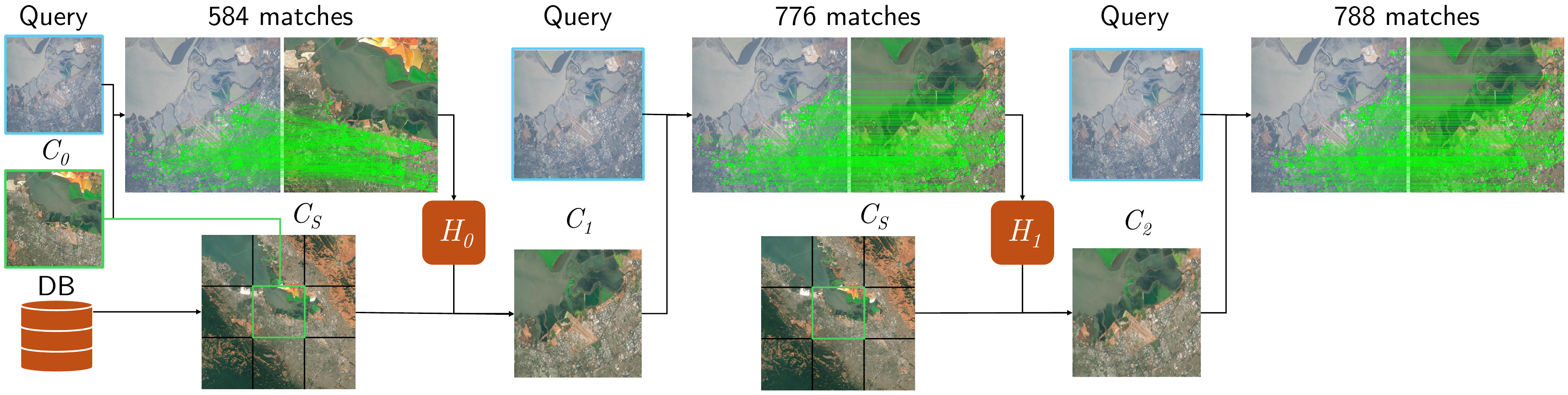}
    \end{center}
    \caption{\textbf{Examples of images during the EarthMatch procedure.} The procedure starts with matching the query $Q$ and the candidate $C_0$ to produce a first homography $H_0$. The candidate surroundings $C_S$ is generated and $H_0$ applied to produce $C_1$. $C_1$ is then matched with $Q$, producing $H_1$ which applied to $C_S$ yields $C_2$. This iterative process continues for a fixed number of iterations (4 iterations in our experiments).
    }
    \label{fig:pipeline_examples_images}
\end{figure*}

Once $C_1$ is generated from $H_0*C_S$, we can obtain a new homography estimate $H_1$, which maps $C_1$ onto $Q$. Since we seek to geolocate with respect to the original candidate, we must track the estimated transformation between $C_0$ to $Q$. This can be computed as $H_Q^1 = H_1*H_0$, as linear transformations can be equivalently applied sequentially by multiplying them.
We repeat this process for a number of iterations, and at each iteration $i$ we have 
$H_Q^i = \prod_{j=0}^{j=i}{H_j}$.
Examples of $Q, C_S, C_0, C_1, C_2$ are shown in \cref{fig:pipeline_examples_images}.

This iterative coregistration process is designed to stop if the matcher $M$ predicts no overlap between the query and the candidate at any iteration of the pipeline. In such cases, EarthMatch will restart the algorithm with the next retrieved image from EarthLoc.
In practice, we run EarthMatch for at most four iterations. At any iteration, the loop stops if the prediction is deemed invalid due to any of the following criteria: (i) the number of matches is less than 4 (too few to compute a homography), (ii) the predicted footprint is non-convex, indicating a failure in matching, or (iii) the predicted footprint has area bigger than $C_S$, \ie 9 times larger than $C_0$, again indicating significant misalignment from matching.
These stopping criteria ensure that almost no false positives are predicted, \ie when the iterative process has successfully finished for a query-candidate pair, it is very unlikely that the final prediction is a false positive.
In practice we found that the majority of models (11 of 16, see ~\cref{sec:experiments})
do not generate \textit{any} false positives when these stopping criteria are applied. However, in cases where false positives are obtained, it is necessary to find a technique to identify and remove them.

To this end, we propose computing a threshold $T_{inl}$ on the number of inliers, and using this threshold to discard predictions with less inliers than $T_{inl}$.
To obtain $T_{inl}$, we fit a logistic regression using the number of inliers for false and true positive astronaut photo/candidate image pairs. We set the threshold to a value that guarantees a 99.9\% probability that any prediction with number of inliers above $T_{inl}$ is in fact a true positive (\ie a precision of 99.9\%).
This is in line with the requirements of APL \cite{Stoken_2023_FMAP}, where having a high precision is more important than high recall.

\section{Dataset}
\label{sec:dataset}

We evaluate our method on images from the Astronaut Imagery Matching Subset (AIMS) dataset \cite{Stoken_2023_FMAP}. These images have been manually located by NASA scientists, 
which have labelled each image's centerpoint in the form of geographic coordinates.
These images are a representative subset of the over 4.5 million photographs of Earth taken by astronauts on the ISS. The full collection can be accessed at the 
Gateway to Astronaut Photography of Earth \footnote{\url{https://eol.jsc.nasa.gov/}}.
From the 323 images within AIMS, we removed 55 photographs with high tilt/oblique, as these often contain Earth limb and cannot be fit with a homography. This results in a subset of 268 images for our benchmark, of which some examples are shown in \cref{fig:ex_queries}.

For each of these, we use an enhanced version of EarthLoc ~\cite{Berton_2024_EarthLoc},
representing the state-of-the-art model for Astronaut Photography Localization through image retrieval, to obtain the top-10 most similar images from a worldwide database of satellite geo-tagged images.
This database contains over 13 million images, at different zooms (\ie different meter-per-pixel resolutions), covering the entire landmass of the planet between latitude 60° and -60° (\ie within the boundaries of the ISS's orbit). This covers all areas that could be depicted in an astronaut photo.
Following EarthLoc \cite{Berton_2024_EarthLoc} we apply $4x$ test-time augmentation on the database, by creating four copies of each database image, one each from a rotation of the original image of 0°, 90°, 180° and 270° respectively.
This not only improves the results of the retrieval stage, but also embeds the top-10 candidates with a coarse estimate of the rotation, with respect to North, of the query: as an example, an ideally perfect retrieval method would find as first prediction a candidate that is no further than 45° rotated from the query.

Even with state-of-the-art retrieval, some queries do not have any positive within the top-10 candidates. Yet, we want our benchmark to be as realistic as possible (reflecting the deployment-time situation of localizing astronaut photos), so we still include these images within the test set.
Of the 268 queries, 244 (91\%) have at least one positive prediction within the top-10 candidates. For EarthMatch, this is the maximum number that can be precisely localized.

\section{Experiments}
\label{sec:experiments}

\subsection{Experimental setup}
EarthMatch is performed on the top-10 retrieval predictions, and can be executed with virtually any image matching technique.
We run the process (matching+homography) for 4 iterations per retrieval candidate, and at each iteration check the stopping criteria to reject invalid predictions.
If the process reaches the end of the 4 iterations, a prediction is obtained, otherwise the process is performed on the next candidate.
For some queries, no prediction is obtained at the end of the iterative process over the top-10 candidates: this characteristic reflects the requirement of APL, for which a prediction should be generated only if it's almost certainly correct, otherwise no prediction should be created.
We compute the homography with a vanilla RANSAC, with the default parameters from OpenCV \cite{itseez2015opencv}.

For our benchmark, we define a simple metric below, and we perform a large number of experiments with different matchers, image resolutions and number of keypoints, as described in the following section.

\myparagraph{Metric}
Each query image is labelled with a manually annotated centerpoint, so we deem an image to be correctly localized if the estimated footprint contains the centerpoint. This is a different metric from other astronaut photography works~\cite{Stoken_2023_FMAP, Bokman_2023_steerers, Berton_2024_EarthLoc}, but is more closely aligned with the localization task (see \cref{sec: apl_metrics} for more details). 
Note that while in theory, this metric leaves open the possibility that a prediction could be considered correct in the case when an inaccurate footprint happens to contain the centerpoint, we find (after visually analyzing hundreds of predictions) that in practice this does not happen: \ie a prediction which contains the centerpoint has almost perfect overlap with the query, as visually shown in \cref{fig:pipeline_examples_images}.

\begin{table*}
\begin{center}
\begin{adjustbox}{width=\textwidth}
\begin{tabular}{l|c|cccc|cc|cc|cc}
\toprule
\multirow{3}{*}{\textbf{Method}} & \multirow{3}{*}{\begin{tabular}[c]{@{}c@{}}\textbf{All} \\ (268 / 244) \end{tabular}} & \multicolumn{4}{c|}{\textbf{Focal Length \textit{f} (mm)}}  & \multicolumn{2}{c|}{\textbf{Camera Tilt}} & \multicolumn{2}{c|}{\textbf{Cloud cover}} & \multirow{3}{*}{\begin{tabular}[c]{@{}c@{}}\textbf{Time} \\ (\emph{sec})\end{tabular}} & \multirow{3}{*}{\begin{tabular}[c]{@{}c@{}}\textbf{Threshold $T_{inl}$} \\ (\emph{\#inliers})\end{tabular}} \\ 
\multirow{2}{*}{}&  & \textit{f} $\le$ 200 \rule{0pt}{0.4cm} & 200 $<$ \textit{f} $\le$ 400 & 400 $<$ \textit{f} $\le$ 800 & \textit{f} $>$ 800 & $<$ 40$^{\circ}$ & $\ge$ 40$^{\circ}$ & $<$ 40 \% & $\ge$ 40 \%  &  &  \\
&  & (90 / 82) & (62 / 55) & (61 / 56) & (55 / 51) & (201 / 184) &  (67 / 60) & (177 / 167) & (91 /  77) & & \\
\midrule
~\textit{Detector-based} & & & & & & & & & \\
SIFT - NN              & 70.5 & 64.6 & 74.5 & 64.3 & 82.4 & 74.5 & 58.3 & 76.6 & 57.1 &    9 &   16 \\                  
ORB - NN               & 23.4 & 23.2 & 23.6 & 17.9 & 29.4 & 26.1 & 15.0 & 30.5 &  7.8 &    6 &   - \\
D2-Net                 & 62.7 & 67.1 & 58.2 & 55.4 & 68.6 & 63.0 & 61.7 & 65.9 & 55.8 &    9 &   16 \\
R2D2                   & 30.7 & 32.9 & 30.9 & 26.8 & 31.4 & 28.3 & 38.3 & 31.7 & 28.6 &    9 &   48 \\
SuperPoint - SuperGlue              & 84.0 & 79.3 & 89.1 & 82.1 & 88.2 & 84.8 & 81.7 & 86.8 & 77.9 &    6 &   - \\
DeDoDe                 & 58.2 & 53.7 & 52.7 & 55.4 & 74.5 & 61.4 & 48.3 & 62.9 & 48.1 &    6 &   - \\
Steerers               & \underline{90.6} & \underline{86.6} & \underline{90.9} & 87.5 & \textbf{100.0} & \textbf{93.5} & 81.7 & \underline{94.6} & \underline{81.8} &    7 &   - \\
ALIKED - LightGlue     & 84.4 & 81.7 & \underline{90.9} & 75.0 & 92.2 & 85.3 & 81.7 & 88.0 & 76.6 &    4 &   - \\
DISK - LightGlue       & 55.3 & 58.5 & 52.7 & 44.6 & 64.7 & 57.1 & 50.0 & 61.1 & 42.9 &    4 &   - \\
DogHardNet - LightGlue & 89.3 & 84.1 & \textbf{96.4} & 87.5 & 92.2 & 90.2 & \underline{86.7} & 93.4 & 80.5 &   12 &   23 \\
SIFT - LightGlue       & 89.3 & 85.4 & 89.1 & \textbf{92.9} & 92.2 & 90.8 & 85.0 & 94.0 & 79.2 &   11 &   - \\
SuperPoint - LightGlue & 78.7 & 73.2 & 81.8 & 78.6 & 84.3 & 79.3 & 76.7 & 81.4 & 72.7 &    4 &   - \\
\midrule
~\textit{Detector-free} & & & & & & & & & \\
Patch2Pix              & 68.0 & 65.9 & 74.5 & 60.7 & 72.5 & 69.0 & 65.0 & 69.5 & 64.9 &    7 &   74 \\
Patch2Pix - SuperGlue  & 83.6 & 79.3 & 89.1 & 82.1 & 86.3 & 84.8 & 80.0 & 86.8 & 76.6 &    7 &   - \\
LoFTR                  & 72.1 & 72.0 & 78.2 & 60.7 & 78.4 & 71.7 & 73.3 & 71.9 & 72.7 &    5 &  305 \\
\midrule
~\textit{Dense matcher} & & & & & & & & & \\
RoMa                   & \textbf{93.0} & \textbf{90.2} & \underline{90.9} & \textbf{92.9} & \textbf{100.0} & \underline{92.4} & \textbf{95.0} & \textbf{95.2} & \textbf{88.3} &   11 &  128 \\
\midrule
Average  & 70.9 & 68.6 & 	72.7 & 	66.5 &77.3&	72.0	 & 67.4	&74.4 &	63.22	&7.31 & -\\
\bottomrule
\end{tabular}
\end{adjustbox}
\end{center}
\vspace{-3mm}
\caption{\textbf{Percent of images correctly localized by each model, for different subsets of AIMS.}
Each column indicates a different subset of AIMS, with the subset depending on the image acquisition conditions - camera focal length, camera tilt (\ie the obliqueness of the camera w.r.t. Earth's surface) and the cloud coverage in the image.
In each column header, the two numbers indicate the number of queries within the subset and the number of localizable queries, \ie those for which at least one of the top-10 retrieval predictions is correct.
Results show the \textbf{percentage of queries correctly localized among the ``localizable'' ones}, so that the upper bound is 100\%.
Methods with threshold $T_{inl} = -$ do not produce any false positives - negative candidates are rejected before 4 iterations - so there is no threshold to compute.
Best \textbf{bolded}, second best \underline{underlined}.
Time is seconds per astronaut photo query for evaluation of all 10 candidates per query. Each model's score is shown for best \emph{image size, number of keypoints} configuration (see \cref{tab:t2}).
}
\label{tab:main_table}
\end{table*}

\begin{table*}
\begin{center}
\begin{adjustbox}{width=\textwidth}
\begin{tabular}{l|ccccccccccccccccccccccc}
\toprule
Model & SIFT - NN & ORB - NN & D2-Net & R2D2 & SuperPoint - SG & DeDoDe & Steerers & ALIKED - LG & DISK - LG & DogHardNet - LG & SIFT - LG & SuperPoint - LG & Patch2Pix & Patch2Pix - SG & LoFTR & RoMa \\
Best \# Keypoints  & 8192 & 8192 & - & 8192 & 2048 & 4096 & 4096 & 4096 & 2048 & 8192 & 4096 & 1024 & - & 2048 & - & 1024 \\
Best Image Size  & 768 & 1024 & 768 & 1024 & 768 & 768 & 768 & 1024 & 768 & 512 & 256 & 768 & 768 & 768 & 768 & 1024 \\
\bottomrule
\end{tabular}
\end{adjustbox}
\end{center}
\vspace{-3mm}
\caption{\textbf{Best image size and number of keypoints for each model}, resulting from a grid search for each method with exponentially growing image sizes between $64 \times 64$ and $1024 \times 1024$, and number of keypoints of 1024, 2048 and 4096. These are the hyperparameters used in \cref{tab:main_table}.
D2-Net, Patch2Pix and LoFTR do not take as input the number of keypoints. SP stands for SuperGlue, LG for LightGlue.}
\vspace{-3mm}
\label{tab:t2}
\end{table*}

\myparagraph{Image Size} We repeat the main experiment with input image sizes ranging from 64 to 1024 px width/height (images are square), resized with a naive resize operation (no padding/cropping).
The same size is applied to both the query and candidates.

\myparagraph{Number of Keypoints} When a matcher is parameterized by the maximum number of keypoints, we analyze how matching performance changes as we adjust this value. We test max keypoints in (1024, 2048, 4096, 8192).

\begin{figure}
    \begin{center}
    \includegraphics[width=0.99\linewidth]{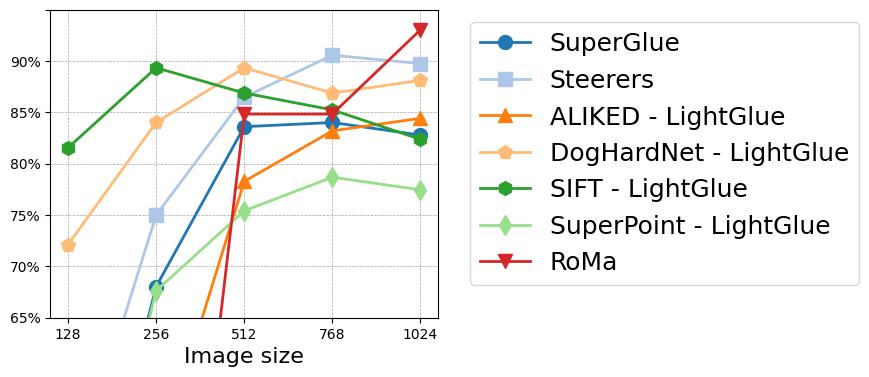}
    \end{center}
    \vspace{-3mm}
    \caption{\textbf{Changing resolution.} Although RoMa is best overall, SIFT-LightGlue is best with low-resolution images, followed by DogHardNet-LightGlue.
    All learnable descriptors rapidly fail as images' resolution decreases.
    }
    \vspace{-3mm}
    \label{fig:plots_resolution}
\end{figure}

\begin{figure}
    \begin{center}
    \includegraphics[width=0.99\linewidth]{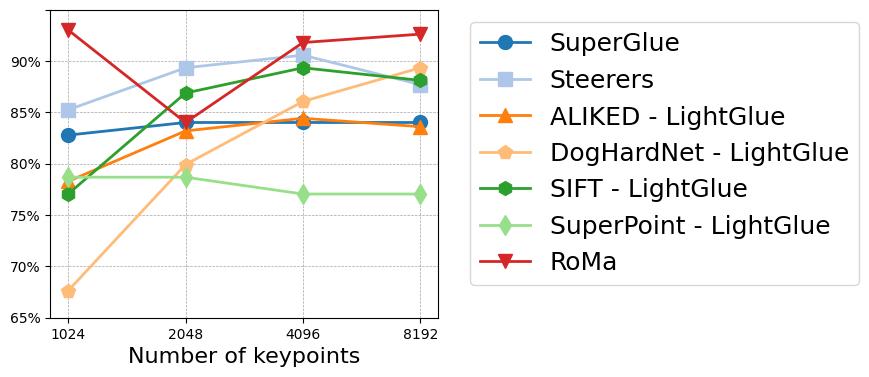}
    \end{center}
    \vspace{-3mm}
    \caption{\textbf{Changing the number of keypoints} for the best-performing models. A flat line (only LoFTR) indicates that the model does not take as input the number of keypoints.
    }
    \vspace{-5mm}
    \label{fig:plots_kpts}
\end{figure}

\myparagraph{Matchers}
We conduct a comprehensive benchmark of matchers, inserting matching models from different families of methods. We divide models into three main families: (i) detector-based local feature descriptors, with either nearest neighbor (NN) matching or learned matchers (\eg SuperGlue~\cite{Sarlin_2020_superglue} or LightGlue~\cite{lindenberger2023lightglue}); (ii) detector-free matchers, and finally (iii) dense matchers, which estimate a dense warp between two images.
Specifically, we test:
\begin{itemize}
    \item \textbf{Detector-based}: we consider handcrafted methods (SIFT~\cite{Lowe_2004_sift}, ORB~\cite{artal_2015_orb}) as well as more recent learned features (Dog-AffNet-HardNet~\cite{Mishkin_2018_repeatability}, D2Net~\cite{Dusmanu_2019_D2Net}, R2D2~\cite{Revaud_2019_r2d2}, DISK~\cite{Tyszkiewicz_2020_disk}, ALIKED~\cite{Zhao_2023_aliked}, DeDoDe~\cite{Edstedt_2023_dedode}). Among these, R2D2 and D2Net require multiple forward passes with increasing resolution up to a user specified threshold. Steerers~\cite{Bokman_2023_steerers} was recently proposed to be rotation invariant. In addition to NN matching, we combine them with LightGlue~\cite{lindenberger2023lightglue} where possible;
    \item \textbf{Detector-free}: among these, LoFTR~\cite{Sun_2021_loftr} is the most popular. We also experiment with Patch2Pix~\cite{Zhou_2021_patch2pix}, which can also be utilized as a match refiner on top of SuperGlue~\cite{Sarlin_2020_superglue};
    \item \textbf{Dense matching}: we use RoMa~\cite{Edstedt_2023_roma}, which recently generated community interest, showing convincing results on wide baseline matching benchmarks. Although it does not directly output keypoints, they can be sampled in an arbitrary quantity, and thus we treat number of keypoints as a hyperparameter, similar to detector-based methods.
\end{itemize}

\subsection{Results}

\label{sec:results}
\myparagraph{Image Size}
In \cref{fig:plots_resolution} we observe varying performance depending on image size.
Most models see performance degradation as image size decreases below 512 px, but the highest performing model at higher resolution, RoMA, sees the steepest dropoff, with no successful matches at 256 px sizes and lower. Dense models like RoMa appear to have a minimum size for matchability, whereas other sparse models have a more gradual performance dropoff.
In the increasing size direction, many models begin to plateau at 512 px. This suggests that for these models, matching on larger images does not improve performance, only increases runtime/compute.

\myparagraph{Number of Keypoints} There are two patterns that emerge from the (maximum) number of keypoints experiments (\cref{fig:plots_kpts}). For most models, performance increases with number of keypoints, plateauing somewhere between 2048 and 4096.
This generally matches expectations, as more keypoints in an image increase the chances the same keypoint is selected in both images, increasing the change of forming a good correspondence.

\myparagraph{AIMS Subsets} 
In \cref{tab:main_table} we examine matching performance on multiple subsets of AIMS, particularly with different focal lengths, low/high cloud cover (occlusion) and low/high tilt (\ie the angle at which the camera was held, inducing larger perspective change in query images). Generally, RoMa is the best performing model across scenarios, showcasing robustness under strong visual changes. Due to its dense matching nature it requires a larger number of inliers for identifying good predictions. Among detector-based matchers, Steerers is the clear winner, achieving good performance with a much lower number of inliers, and at almost twice the speed of RoMa.
In general, the methods of ALIKED + LightGlue, Steerers and RoMa produce optimal speed-accuracy trade-offs.

While the majority of models do not produce any false positives (\ie for wrong candidates, EarthMatch stops before reaching the final iteration, thus rejecting the candidate), a few models do admit them, requiring computation of an inlier threshold $T_{inl}$ to further validate correctness. When a threshold is computed, candidates must exceed this value to be considered a confident prediction.
$T_{inl}$ can be strongly affected by even a single highly confident false positive. This can increase $T_{inl}$ such that a number of true positives are discarded as their inlier count does not exceed this inflated $T_{inl}$.
This behaviour, seen in the rightmost column of \cref{tab:main_table}, explains the uneven behavior of the RoMa curve in \cref{fig:plots_kpts} and \cref{fig:plots_resolution}, where just a single high-confident false positive causes a strong dip in results.

In analyzing different subsets of AIMS, the most challenging setting is high cloud cover, which creates a natural occlusion to the matching process. Most methods also benefit from low camera tilt, which leads to high similarity between the query and the candidate. RoMa presents a notable exception to this trend, achieving better performances with high camera tilt.

\subsection{Limitations}

\myparagraph{Limited hyperparamer tuning}
One of the goals of this benchmark is the fair evaluation of existing image matching models on the out-of-distribution (w.r.t. their training sets) domain of Astronaut Photography Localization.
To this end, in our evaluation we tuned only the two hyperparameters that are recurrent across multiple models, namely the input image size and the number of keypoints.
For fair evaluations, we decided not to tune any model-specific hyperparameters, given that
(1) this would give an advantage to methods with more hyperparameters and (2) it would lead to an explosion in the number of experiments.
We therefore note that despite EarthMatch showing the potential of each model on the domain of APL, these results could potentially still be improved with per-model hyperparameter optimization, including the threshold used for RANSAC, which could be an interesting future development.

\myparagraph{Map Projection}
Finally, in this work we used a map projected representation of Earth. In particular, we used the Mercator projection, which preserves angles but not areas. While this can be problematic when viewing large areas (\ie visually, areas closer to the poles appear larger than they are), 
we found this to have no noticeable impact on matchability or the estimated footprint of considered images, due to the images having a relatively small area with respect to the Earth's surface. It's possible other projections could yield improved matching in some, likely more polar, regions where the Mercator projection is visually more dissimilar to the true view of the Earth. 
This would have a stronger impact when registering Earth limb photos, which are however not considered within our benchmark.

\section{Conclusion}
\label{sec:conclusion}

In this work we present a pipeline to reliably and confidently estimate the footprint of astronaut photographs.
The pipeline is made of a pre-existing retrieval method \cite{Berton_2024_EarthLoc} and our newly introduced EarthMatch, an iterative coregistration algorithm that takes advantage of image matching models to output a predicted footprint and confidence.
We run a large number of experiments with many matchers using different images sizes and number of keypoints, thoroughly evaluating their usability in the domain of Astronaut Photography Localization.

To foster future research and simplify reproducibility, we release the post-retrieval dataset of astronaut photo query and top 10 candidates used for our experiments, relieving researchers to having to compute the time-consuming large-scale retrieval step. We also release the code to replicate all experiments within this paper. This code is trivially extensible to future matching methods or other domains.

Finally, our pipeline offers an efficient and robust method for astronaut photography localization that can immediately be deployed on the existing 4.5 million and growing database of astronaut photos of Earth.

\myparagraph{Acknowledgements}
\small{We acknowledge the CINECA award under the ISCRA initiative, for the availability of high performance computing resources.
This work was supported by CINI. G. Goletto is supported by PON “Ricerca e Innovazione” 2014-2020 – DM 1061/2021 funds.
Project supported by ESA Network of Resources Initiative.
This study was carried out within the project FAIR - Future Artificial Intelligence Research - and received funding from the European Union Next-GenerationEU (PIANO NAZIONALE DI RIPRESA E RESILIENZA (PNRR) – MISSIONE 4 COMPONENTE 2, INVESTIMENTO 1.3 – D.D. 1555 11/10/2022, PE00000013). This manuscript reflects only the authors’ views and opinions, neither the European Union nor the European Commission can be considered responsible for them.
European Lighthouse on Secure and Safe AI – ELSA, HORIZON EU Grant ID: 101070617
}

{
    \small
    \bibliographystyle{ieeenat_fullname}
    \bibliography{main}
}

% % WARNING: do not forget to delete the supplementary pages from your submission 
\clearpage
\setcounter{page}{1}
\maketitlesupplementary

\section{Astronaut Photography Localization Metrics}
\label{sec: apl_metrics}
Since the introduction of the Astronaut Image Matching Subset (AIMS) dataset in FMAP~\cite{Stoken_2023_FMAP}, various metrics have been proposed to best capture the astronaut photography localization challenge. FMAP itself uses a fixed set of reference images provided as part of AIMS, and for each astronaut photograph, there may be one or more matching reference images. FMAP measures average precision over these reference images.  Such a metric emphasizes high recall matching, such that any query/reference pair with sufficiently overlapping extents should match.

FMAP uses a fixed, discrete set of inlier thresholds and at most 100 negatives in it's average precision calculation. Steerers~\cite{Bokman_2023_steerers} expands this calculation to a continuous inlier range, and uses all of the negatives in AIMS. This gives a more complete, but otherwise comparable, metric. 

EarthLoc~\cite{Berton_2024_EarthLoc} introduces the concept of retrieval to astronaut photography localization, and brings a retrieval oriented metric to the task: Recall@N. This is the percent of astronaut photos in which one of the top N retrieved reference images is correct, with correctness defined as having non-zero overlap between the astronaut photo and reference image. Due to recasting APL as a retrieval, rather than a pairwise matching task, this metric is not directly comparable to the average precision of FMAP and Steerers. 

In this work, we again shift the metric, this time to more closely align it with the downstream localization task. Since the end goal of APL is to find the correct location of an astronaut photograph on Earth, and our clearest indication of the correct place on Earth is the manually annotated center point, we use the number (or, equivanetly, percent) of photos where our predicted footprint contains the centerpoint as a metric, and denote this as ``images correctly localized". This is a better measure of real world performance and allows us to compare methods as they would operate in deployment.

\section{Qualitative results}
\label{sec: quali}
\begin{figure*}
    \begin{center}
    \includegraphics[width=\linewidth]{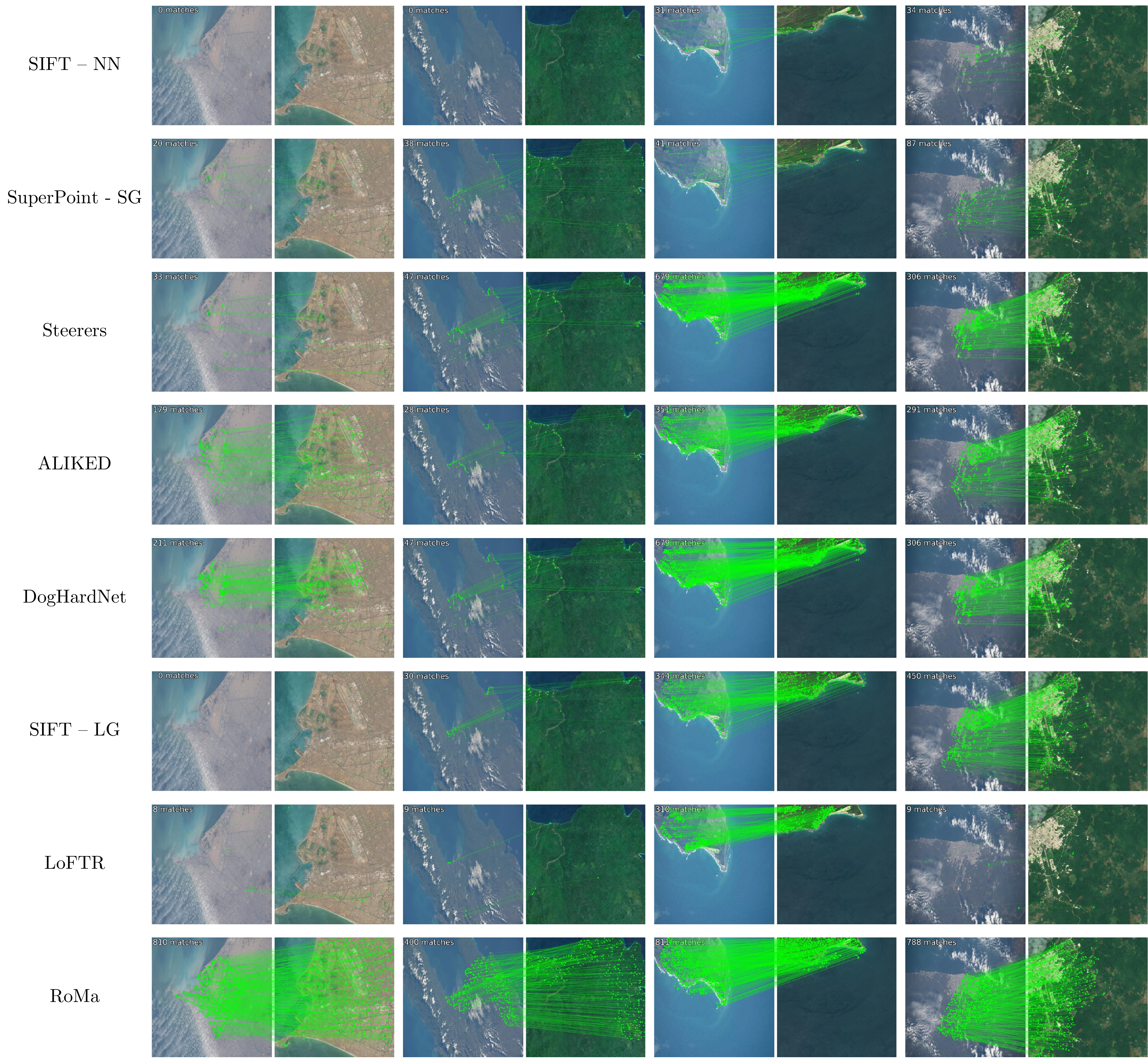}
    \end{center}
    \caption{\textbf{Qualitative results} for a select number of matching methods proposed in our benchmark. 
    }
    \label{fig:qualitative}
\end{figure*}

In \cref{fig:qualitative} we show examples of matchings with different models.

\end{document}